\documentclass[letterpaper,journal]{IEEEtran}
\usepackage{amsmath,amsfonts}
\usepackage{array}
\usepackage[caption=false,font=normalsize,labelfont=sf,textfont=sf]{subfig}
\usepackage{textcomp}
\usepackage{stfloats}
\usepackage{float}
\usepackage{url}
\usepackage{verbatim}
\usepackage{graphicx}
\usepackage{booktabs}
\usepackage{cite}
\newcommand{\meanstd}[2]{\ensuremath{#1\,\pm\,#2}}
\newcommand{\boldmeanstd}[2]{\ensuremath{\boldsymbol{#1\,\pm\,#2}}}
\title{HC2Robot: Hyper-Connection Transformers with Predicted Foresight for Multi-Terrain Humanoid Parkour}
\author{Wenhao Xu$^{*}$, Yanheng Mai$^{*}$, Yifei Fu$^{*}$, Nan Zheng, Yanbo Zhou, Zirui Huang, \\
Mingsheng Ling, Xinjue Wang, and Renjing Xu$^{\dagger}$%
\thanks{$^{*}$Wenhao Xu, Yanheng Mai and Yifei Fu contributed equally. $^{\dagger}$Corresponding author: Renjing Xu.}%
\thanks{Wenhao Xu, Yanheng Mai, and Renjing Xu are with The Hong Kong University of Science and Technology (Guangzhou) (e-mail: renjingxu@hkust-gz.edu.cn).}%
\thanks{Wenhao Xu, Yanheng Mai, Yifei Fu, Nan Zheng, Yanbo Zhou, Zirui Huang, Mingsheng Ling, and Xinjue Wang are with CLAI-LAB, CL-TECH.}%
\thanks{Yifei Fu and Zirui Huang are with South China Agricultural University.}%
\thanks{Mingsheng Ling is with Southern University of Science and Technology.}%
}
\begin{document}
\maketitle

\begin{abstract}
Humanoid parkour requires robots to reason about future footholds and whole-body motions over highly discontinuous terrain. We revisit the information asymmetry in asymmetric actor-critic learning for perceptive humanoid locomotion: the critic exploits privileged simulator signals during training, whereas the deployable actor operates solely from onboard perception. We argue that the key is not exposing privileged information to the actor, but learning a deployable approximation of it to narrow the information gap between training and deployment. We present HC2Robot, a deployable perceptive humanoid parkour framework consisting of a Predicted Foresight Module (PFM) and a Hyper-Connection (HC) Transformer. PFM predicts from deployable observations the policy-conditioned timing, location, and support quality of each foot's next touchdown event, converting privileged supervision into a deployable latent representation that provides the actor with predicted touchdown information without privileged runtime sensing. The HC Transformer further enhances temporal reasoning through learnable multi-stream routing, while a future motion prediction head provides short-horizon motion priors for regularization. Across stairs, gaps, high platforms, and slopes, HC2Robot achieves a mean simulated traversal success rate of 97.60\% and consistently higher safe foothold rates than strong perceptive locomotion baselines, supporting improved traversal performance and geometric foothold support under deployable sensing.
\end{abstract}


\section{Introduction}

\IEEEPARstart{H}{umanoid} parkour demands traversal of stairs, gaps, platforms, and slopes while maintaining dynamically feasible whole-body motion \cite{zhuang2024humanoid,wu2026perceptive}. Success requires the controller to coordinate perception, contact timing, and body motion well before each foothold is realized. While massively parallel reinforcement learning has enabled robust locomotion \cite{hwangbo2019agile,rudin2022learning}, proprioception-only policies lack foresight of upcoming terrain discontinuities \cite{lee2022learning,nahrendra2023dreamwaq}. Depth-conditioned perception extends this horizon with geometric priors \cite{agarwal2023egocentric,cheng2023extreme}, enabling scalable perceptive parkour over challenging terrain \cite{zhu2026hiking,wu2026perceptive}.

Two challenges remain for deployable perceptive parkour. First, temporal representations must effectively integrate proprioceptive and depth-conditioned information. Conventional residual Transformers use a single residual stream \cite{he2016resnet,vaswani2017attention}; multi-stream routing offers additional paths for cross-layer feature mixing, complementing recent advances in cross-modal fusion, mixture-of-experts, and reference-gait priors \cite{hao2026cref,huang2026core,qin2026prior}. Second, simulator-derived contact and support signals enhance learning but are unavailable at deployment \cite{pinto2018asymmetric,kumar2021rma,lee2022learning}. We train an auxiliary estimator with simulator-derived labels to make task-relevant predictions available to the actor.

HC2Robot addresses these challenges through two complementary mechanisms. A Hyper-Connection (HC) Transformer \cite{Zhu2024HyperConnections} replaces the conventional single residual pathway with learnable multi-stream aggregation and mixing, enabling richer cross-layer information exchange while evaluating each attention and feed-forward sublayer only once. A Predicted Foresight Module (PFM), trained with asymmetric actor–critic learning \cite{pinto2018asymmetric,lambrechts2025theoreticaljustificationasymmetricactorcritic}, forecasts each foot's next touchdown event---its time-to-contact, touchdown location, and support quality---from deployable actor features and fuses the forecast into the policy. Additionally, an auxiliary motion predictor adopted from ParkourFormer \cite{mai2026parkourformer} provides two-step future AMP states for motion regularization \cite{Peng2021AMP}. All prediction pathways remain deployment-compatible, using simulator information solely as training supervision.

\begin{figure*}[!t]
\centering
\includegraphics[trim={6.02249pt 2.00750pt 4.01499pt 25.09369pt},clip,width=0.99199359\textwidth]{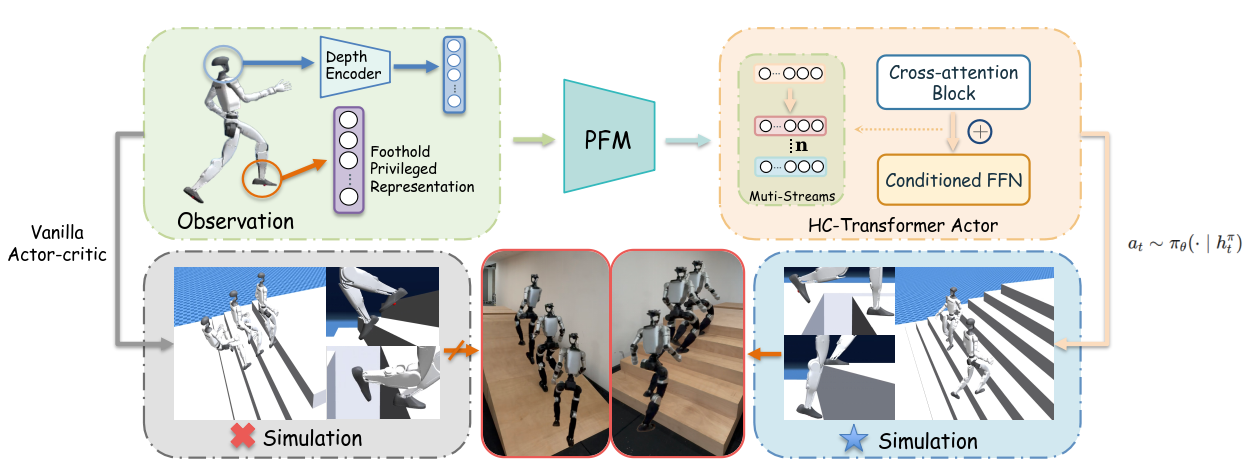}
\caption{HC2Robot overview. Simulator foothold descriptors supervise PFM during training; deployment uses predictions from proprioception and depth, fused at the policy head. Lower panels show simulated traversal comparisons and physical deployment.}
\label{fig:HC2Robot_overview}
\end{figure*}

We evaluate HC2Robot on stairs, gaps, high platforms, and slopes; the results demonstrate that combining HC feature reuse with PFM foothold foresight consistently improves traversal success and safe foothold rate over strong baselines.
Fig.~\ref{fig:HC2Robot_overview} summarizes the framework.

Our contributions are as follows:
\begin{itemize}
\item A Hyper-Connection Transformer that replaces standard residual connections with learnable multi-stream aggregation and mixing, enabling richer temporal feature reuse while evaluating each attention and feed-forward sublayer only once.
\item A Predicted Foresight Module (PFM) that forecasts the timing, location, and support quality of each foot's next touchdown event through a continuous time-to-event formulation, providing supervised foothold foresight without a fixed prediction offset or privileged signals in the deployment interface.
\item Simulation results across four challenging terrain classes demonstrating that HC2Robot consistently improves traversal success and safe foothold rate over strong perceptive parkour baselines.
\end{itemize}
\section{Related Work}

\subsection{Perceptive Locomotion and Humanoid Parkour} Massively parallel reinforcement learning supports robust blind locomotion \cite{hwangbo2019agile,rudin2022learning}, while depth- and vision-conditioned policies extend it to terrain that proprioception alone cannot resolve \cite{lee2022learning,agarwal2023egocentric,nahrendra2023dreamwaq}. Perceptive parkour has since been demonstrated on quadrupeds \cite{zhuang2023robot,cheng2023extreme,hoeller2023anymal} and humanoids \cite{zhuang2024humanoid}, with deployable sequence control and multimodal terrain fusion becoming common design choices \cite{radosavovic2023realworld,li2024learning,zhu2026hiking,wu2026perceptive,crismariu2026march,hao2026cref,huang2026core,qin2026prior}. HC2Robot adopts this deployable-perception setting but focuses on using a predicted, task-relevant foothold event to narrow the training--deployment information gap; its HC backbone and PFM complement these locomotion and parkour systems.

\subsection{Privileged-State Methods and Learning Paradigms}

Asymmetric actor--critic learning separates privileged critic inputs from the deployable policy \cite{pinto2018asymmetric,lambrechts2025theoreticaljustificationasymmetricactorcritic}. Lee et al. \cite{lee2022learning} and RMA \cite{kumar2021rma} use deployable observations to support terrain adaptation or latent dynamics estimation, while concurrent estimator--controller training learns hidden robot quantities jointly \cite{ji2022concurrent}; teacher--student and latent-scaffold methods transfer privileged actions or representations \cite{zhang2025distillation,jin2025teacher,wu2025learntoteach,ma2026vpies}. These approaches estimate current state/dynamics or imitate a privileged policy, whereas privileged-state rewards only shape optimization and do not expose a deployable representation. PFM instead uses simulator descriptors solely as training labels: from proprioception and depth, it predicts the next policy-conditioned touchdown time, position, support attributes, and uncertainty, then fuses that event forecast without privileged actor inputs or teacher actions.

\subsection{Foothold Reasoning, Predictive Supervision, and Motion Priors}

\begin{figure*}[!t]
\centering
\includegraphics[trim={4.41649pt 2.40899pt 2.40899pt 4.41649pt},clip,width=0.99291667\textwidth]{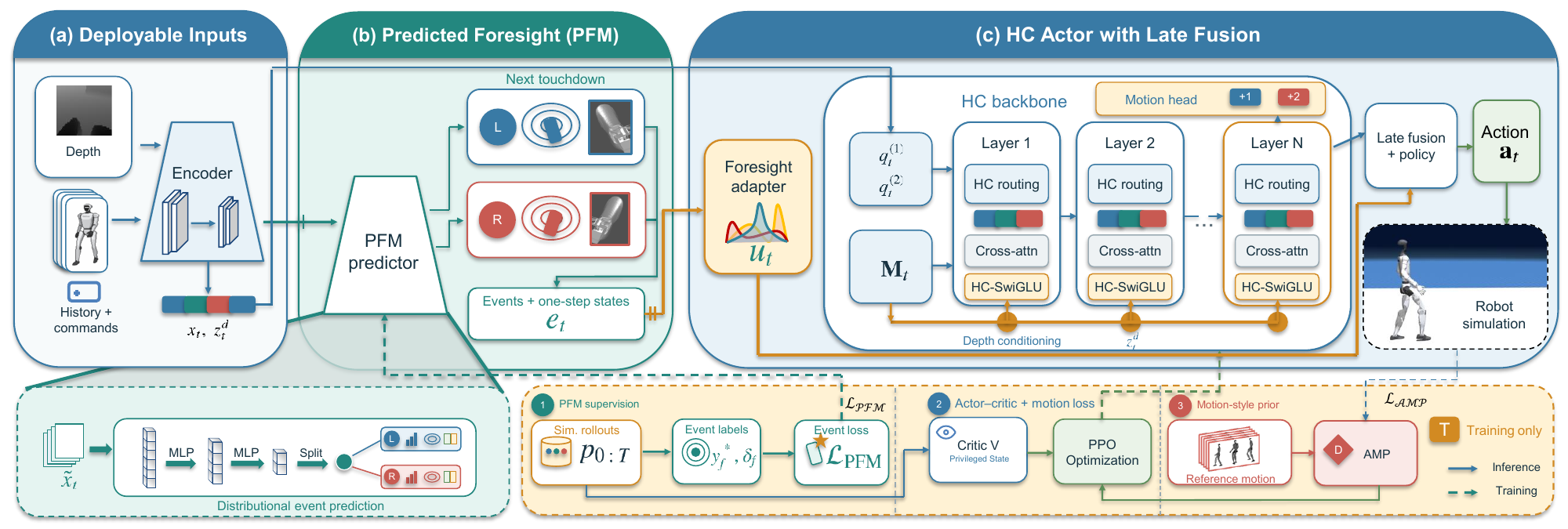}
\caption{HC2Robot architecture. (a) Observation encoding from depth and proprioceptive histories. (b) Predicted Foresight Module (PFM), forecasting each foot's next touchdown timing, location, and support quality. (c) Hyper-Connection (HC) Transformer actor with multi-stream temporal routing and late fusion of predicted foresight for action generation.}
\label{fig:HC2Robot_architecture}
\end{figure*}

Terrain-aware perception, terrain imagination, and internal predictive models improve representations of terrain or hidden dynamics \cite{fahmi2022vital,nahrendra2023dreamwaq,long2024perceptive}; their targets are not the next contact event. TAGA \cite{taga2026} predicts a task-relevant terrain region with active gaze, while PFM predicts the next touchdown time, location, support attributes, and uncertainty. ParkourFormer \cite{mai2026parkourformer} combines predictive supervision with humanoid sequence modeling, whereas PFM uses foothold-specific labels and a censoring-aware event objective. AMP supplies a motion-distribution prior \cite{Peng2021AMP}; HC2Robot keeps touchdown-event supervision distinct from motion regularization.
Fig.~\ref{fig:HC2Robot_architecture} shows the architecture.

\section{Method}
\label{sec:method}

HC2Robot combines a Hyper-Connection (HC) Transformer with a Predicted Foresight Module (PFM) under asymmetric actor--critic training \cite{pinto2018asymmetric}: the HC backbone improves temporal feature reuse, while PFM forecasts each foot's next touchdown event from deployable observations.

\subsection{Asymmetric Observation and Foothold Supervision}

We model humanoid parkour as a partially observable control problem. At step $t$, the actor receives an onboard depth image $\mathcal I_t$ and a proprioceptive observation
\begin{equation}
o_t=[\omega_t,\; g_t,\; c_t,\; q_t,\; \dot q_t,\; a_{t-1}],
\label{eq:deployable_obs}
\end{equation}
where $\omega_t\in\mathbb{R}^3$ is the angular velocity in the base coordinate system, $g_t\in\mathbb{R}^3$ is the projected gravity direction in the base coordinate system, $c_t\in\mathbb{R}^3$ is the velocity command, $(q_t,\dot q_t)\in\mathbb{R}^{n_j\times 2}$ are the joint positions and velocities of the humanoid robot with $n_j$ degrees of freedom, and $a_{t-1}\in\mathbb{R}^{n_j}$ is the previous action.

Since contact transitions occur over multiple time steps, the policy retains a history of length $L$, $H_t=(o_{t-L+1},\ldots,o_t)$. The depth encoder $E_d$ compresses the recent depth-image history into a context token $z_t^d=E_d(\mathcal I_{t-L+1},\ldots,\mathcal I_t)\in\mathbb{R}^{D_z}$, where $D_z$ is the dimension of the token. The actor encoder $E_a$ then combines $H_t$ with $z_t^d$ to produce the deployable feature
\begin{equation}
x_t = E_a(H_t,\; z_t^d).
\label{eq:actor_feature}
\end{equation}
The depth encoder architecture follows prior work \cite{mai2026parkourformer}; its parameters are trained end-to-end with the policy.

During training, the critic receives privileged simulator variables that are unavailable at deployment: base linear velocity $v_t\in\mathbb{R}^3$, center-of-mass position $m_t\in\mathbb{R}^3$, and a two-foot support descriptor $p_t\in\mathbb{R}^{14}$. For each foot $f\in\{L,R\}$, the descriptor is defined as
\begin{equation}
p_{t,f}=[b_{t,f},\; \Delta x^b_{t,f},\; \Delta y^b_{t,f},\; r_{t,f},\;
\eta_{t,f},\; \rho_{t,f},\; \phi_{t,f}]\in\mathbb{R}^7,
\label{eq:foothold_descriptor}
\end{equation}
where $b_{t,f}\in\{0,1\}$ indicates whether the resultant contact force exceeds a threshold; $(\Delta x^b_{t,f},\Delta y^b_{t,f})$ are the mean coordinates of nearby ray-hit points expressed in the base frame; $r_{t,f}$ is the support ratio—the fraction of rays terminating on terrain within the foot sole grid; $\eta_{t,f}$ is the terrain risk score measuring local height variance; $\rho_{t,f}$ is the load-weighted support; and $\phi_{t,f}$ is the normalized normal-force proxy. Concatenating both feet gives $p_t=[p_{t,L};p_{t,R}]$, which serves two roles in training: it provides conditioning for the value function, and its temporal trajectory supplies the raw signal from which touchdown-event supervision is constructed for the Predicted Foresight Module (PFM)---contact transitions define the event times, and the descriptor values recorded at those instants define the event attributes (Sec.~\ref{sec:foresight}). The actor never directly observes $(v_t,m_t,p_t)$; its only exteroceptive input is the depth token $z_t^d$. Table~\ref{tab:actor_critic_comparison} compares the interfaces.

\subsection{HC-Transformer Backbone}

The temporal backbone builds on a cross-attention architecture over the deployable history, replacing standard residual connections with Hyper-Connection (HC) routing \cite{Zhu2024HyperConnections}. The encoded proprioceptive history is reshaped into an $L$-step token sequence, which together with the depth token $z_t^d$ forms the key--value memory $M_t$ for cross-attention. Two learned query tokens, conditioned on $z_t^d$, focus on future-relevant temporal dependencies and are replicated across $S$ parallel streams of dimension $D$, forming the multi-stream state $Z\in\mathbb{R}^{S\times 2\times D}$.

Each attention and feed-forward sublayer follows the same three-step routing with independent parameters $(\alpha_r,\beta_r,R_r)$, where $r\in\{\mathrm{attn},\mathrm{ffn}\}$ indexes the sublayer. The streams are first aggregated with learnable coefficients $\alpha_r\in\mathbb{R}^{S}$:
\begin{equation}
\bar Z_r = \sum_{s=1}^{S} \alpha_{r,s} Z_s.
\end{equation}
The sublayer is evaluated once on this aggregated representation: for the attention path,
\begin{equation}
U_{\text{attn}} = \mathcal{A}\!\left(\bar Z_{\text{attn}},\; M_t\right),
\end{equation}
where $\mathcal{A}$ denotes cross-attention with $M_t$ as key and value, and for the feed-forward path,
\begin{equation}
U_{\text{ffn}} = \mathcal{F}\!\left(\bar Z_{\text{ffn}},\; z_t^d\right),
\end{equation}
where $\mathcal{F}$ is a depth-conditioned SwiGLU transformation \cite{shazeer2020glu}. The output is then distributed back to the streams and mixed with their residual pathways:
\begin{equation}
Z'_{r,s} = \text{LN}\!\left(\beta_{r,s}\,U_r + \sum_{j=1}^{S} R_{r,sj}\,Z_j\right),
\label{eq:hc_routing}
\end{equation}
where $\beta_r\in\mathbb{R}^{S}$ scales how each stream receives the update and $R_r\in\mathbb{R}^{S\times S}$ mixes the residual paths for sublayer $r$. Thus, $(\alpha_r,\beta_r,R_r)$ generalize the standard $x+f(x)$ residual by maintaining stream-specific representations while evaluating each sublayer once on $\bar Z_r$.

After $N$ layers, the streams are summed and normalized to produce a single temporal feature for each query token, which serves as the policy input. The critic follows an analogous HC structure over its privileged history, with a separate depth encoder and without the foothold predictor.
\begin{table*}[!t]
\footnotesize
\centering
\caption{Actor--critic interfaces per frame. Actor: deployable observations and PFM forecasts; critic: privileged states. The 22-D forecast contains two 10-D event summaries and two confidence values; $H\!\times\!W$ denotes depth resolution.}
\label{tab:actor_critic_comparison}
\setlength{\tabcolsep}{3pt}
\resizebox{\textwidth}{!}{%
\begin{tabular}{lcccccccc}
\toprule
\textbf{Signal} & \multicolumn{4}{c}{\textbf{Vanilla}} & \multicolumn{4}{c}{\textbf{Ours}} \\
\cmidrule(lr){2-5}\cmidrule(lr){6-9}
& \multicolumn{2}{c}{\textbf{Actor}} & \multicolumn{2}{c}{\textbf{Critic}} &
\multicolumn{2}{c}{\textbf{Actor}} & \multicolumn{2}{c}{\textbf{Critic}} \\
\cmidrule(lr){2-3}\cmidrule(lr){4-5}\cmidrule(lr){6-7}\cmidrule(lr){8-9}
\midrule
Base angular velocity $\omega_t$ & $3$ & IMU & $3$ & IMU & $3$ & IMU & $3$ & IMU \\
Projected gravity $g_t$ & $3$ & IMU & $3$ & IMU & $3$ & IMU & $3$ & IMU \\
Command $c_t$ & $3$ & command input & $3$ & command input & $3$ & command input & $3$ & command input \\
Joint position $q_t$ & $29$ & joint encoders & $29$ & joint encoders & $29$ & joint encoders & $29$ & joint encoders \\
Joint velocity $\dot q_t$ & $29$ & joint encoders & $29$ & joint encoders & $29$ & joint encoders & $29$ & joint encoders \\
Previous action $a_{t-1}$ & $29$ & history & $29$ & history & $29$ & history & $29$ & history \\
Depth history $\mathcal I_t$ & $H\!\times\!W$ & RealSense/RayCaster & $H\!\times\!W$ & RayCaster & $H\!\times\!W$ & RealSense/RayCaster & $H\!\times\!W$ & RayCaster \\
\midrule
\multicolumn{9}{c}{\textit{Privileged training-time signals (critic: simulator values; actor: PFM forecasts)}} \\
\cmidrule(lr){1-9}
Base velocity $v_t$ & -- & -- & $3$ & simulator & $3$ & PFM one-step prediction & $3$ & simulator \\
Center-of-mass position $m_t$ & -- & -- & $3$ & simulator & $3$ & PFM one-step prediction & $3$ & simulator \\
Foothold descriptor $p_t$ & -- & -- & $14$ & simulator & $22$ & PFM event forecast & $14$ & simulator \\
\bottomrule
\end{tabular}
}
\end{table*}


\subsection{Predicted Foresight and Late Fusion}
\label{sec:foresight}

Rather than regressing the privileged state at a fixed future step, PFM forecasts the next touchdown of each foot as an event whose timing is itself a prediction target. All forecasts are produced from the deployable feature $x_t$; let $\tilde x_t$ denote its detached copy.

For foot $f\in\{L,R\}$, the next touchdown is that foot's first strictly future $0\!\rightarrow\!1$ contact transition; if the foot is in stance, a lift-off must occur before the next touchdown is valid. From the simulator trajectory of the foothold descriptor \eqref{eq:foothold_descriptor}, each rollout frame is annotated with the touchdown event
\begin{equation}
y_f^*=(\tau_f^*,\;\boldsymbol p_f^*,\;r_f^*,\;\eta_f^*,\;\rho_f^*,\;\phi_f^*),
\label{eq:event_label}
\end{equation}
where $\tau_f^*>0$ is the time-to-touchdown measured in control periods, $\boldsymbol p_f^*\in\mathbb R^2$ is the touchdown position expressed in the base frame at the prediction time, and $r_f^*$, $\eta_f^*$, $\rho_f^*$, and $\phi_f^*$ are the support ratio, terrain risk, load support, and normal-force proxy recorded at the touchdown instant. If the episode terminates before that touchdown, the frame is right-censored at the episode boundary, with no observed event time or touchdown attributes.

PFM outputs a distributional estimate of this event for each foot,
\begin{equation}
F_\xi(\tilde x_t)_f=\big(\mu_{\tau,f},\,\sigma_{\tau,f},\,\boldsymbol\mu_{p,f},\,\log\boldsymbol\sigma^2_{p,f},\,\hat{\boldsymbol a}_f,\,\log\boldsymbol\sigma^2_{a,f}\big),
\label{eq:pfm_output}
\end{equation}
where $(\mu_{\tau,f},\sigma_{\tau,f})$ parameterize a log-normal distribution over the time-to-touchdown, whose support on $\mathbb R^+$ matches the non-negative nature of event timing; $(\boldsymbol\mu_{p,f},\log\boldsymbol\sigma^2_{p,f})$ define a diagonal Gaussian over the touchdown position; and $\hat{\boldsymbol a}_f=(\hat r_f,\hat\eta_f,\hat\rho_f,\hat\phi_f)$, together with its log-variance vector $\log\boldsymbol\sigma^2_{a,f}$, covers the four touchdown attributes. Two separate heads provide one-step Gaussian estimates of base velocity and center-of-mass position; they are trained as auxiliary supervision that regularizes the PFM encoder, and their estimates are fused as short-horizon kinematic context alongside the event forecast. All log-variances are clipped to $[-5,2]$.

For control, the per-foot forecast is summarized as
\begin{equation}
\hat y_{t,f}=[\tilde\tau_f;\,\sigma_{\tau,f};\,\boldsymbol\mu_{p,f};\,\log\boldsymbol\sigma^2_{p,f};\,\hat{\boldsymbol a}_f]\in\mathbb R^{10},
\label{eq:event_summary}
\end{equation}
where $\tilde\tau_f=\exp(\mu_{\tau,f})$ is the median time-to-touchdown. A confidence $\kappa_{t,f}\in(0,1]$ is an exponential transform of the foot's mean predicted variance, so that larger uncertainty yields lower confidence \cite{kendall2017uncertainties}. Concatenating both feet with the one-step auxiliary estimates gives the foresight feature
\begin{equation}
e_t=[\hat v_{t+1};\,\hat m_{t+1};\,\hat y_{t,L};\,\hat y_{t,R};
\kappa_{t,L};\,\kappa_{t,R}]\in\mathbb R^{28}.
\label{eq:pfm_feature}
\end{equation}

Let $\tilde e_t$ denote a detached copy of the foresight feature. The adapter first maps it to a 32-dimensional latent:
\begin{equation}
u_t=A_\varphi(\tilde e_t)\in\mathbb R^{32}.
\label{eq:foresight_adapter}
\end{equation}
The foresight latent is then fused with the motion representation $b_t$ only at the policy head:
\begin{equation}
h_t^\pi=W_f[b_t;u_t]+b_f\in\mathbb R^{128}.
\label{eq:late_fusion}
\end{equation}
The action head maps $h_t^\pi$ to a 29-dimensional mean $\mu_t=f_\theta(h_t^\pi)$. During PPO training, the action is sampled from the learned diagonal Gaussian policy
\begin{equation}
a_t\sim\mathcal N\!\left(\mu_t,\operatorname{diag}(\sigma_\theta^2)\right).
\label{eq:action_distribution}
\end{equation}
At deployment, the policy uses the deterministic mean $a_t=\mu_t$.

\subsection{Training and Deployment}
\label{sec:training}

The actor--critic parameters $\theta$ (including HC, the motion head, adapter, and fusion layers), PFM parameters $\xi$, and AMP discriminator parameters $\psi$ are updated sequentially with separate objectives. $\mathcal L_{\mathrm{PPO}}(\theta)$ combines the clipped PPO objective \cite{schulman2017ppo} with masked MSE on the two future motion states adopted from ParkourFormer \cite{mai2026parkourformer}.

PFM uses non-terminal rollout transitions labeled by \eqref{eq:event_label}. Let $m_i\in\{0,1\}$ indicate transition validity, $\delta_{i,f}\in\{0,1\}$ indicate touchdown before episode termination, and $T_{i,f}$ be the remaining control periods for a censored frame. With $M=\sum_i m_i$ and $D=\sum_{i,f}m_i\delta_{i,f}$, its loss is
\begin{equation}
\begin{aligned}
\mathcal L_{\mathrm{PFM}}(\xi)={}&-\frac{1}{2M}\sum_{i,f}m_i\log\big[f_\tau(\tau_{i,f}^*)^{\delta_{i,f}}S_\tau(T_{i,f})^{1-\delta_{i,f}}\big]\\
&+\frac{1}{6D}\sum_{i,f,j}m_i\,\delta_{i,f}\,\ell_{i,f,j}
+\frac{\lambda_{\mathrm{aux}}}{6M}\sum_{i,j}m_i\,\ell^{\mathrm{aux}}_{i,j},
\end{aligned}
\label{eq:pfm_loss}
\end{equation}
where $f_\tau$ and $S_\tau$ are the log-normal density and survival function, respectively, so censored frames contribute only survival likelihood to event-time supervision. The six-dimensional target $d_{i,f}$ contains touchdown position and support attributes, and $d_i^{\mathrm{aux}}$ contains one-step base velocity and center-of-mass position; event regression applies only to observed touchdowns. Each regression term $\ell$ uses $\nu+(d-\hat d)^2e^{-\nu}$ with predicted log-variance $\nu$ \cite{kendall2017uncertainties}, twice the Gaussian negative log-likelihood up to an additive constant; $\lambda_{\rm aux}$ is defined under this scaling.

The AMP discriminator distinguishes detached policy trajectories $\bar\tau^\pi$ from reference trajectories $\tau^r$ using the least-squares loss \cite{Peng2021AMP},
\begin{equation}
\begin{aligned}
\mathcal L_{\mathrm{AMP}}(\psi)={}&\tfrac12\mathbb E_{\bar\tau^\pi}[(D_\psi(\bar\tau^\pi)+1)^2]\\
&+\tfrac12\mathbb E_{\tau^r}[(D_\psi(\tau^r)-1)^2]+\lambda_R\mathcal R(\psi),
\end{aligned}
\label{eq:amp_loss}
\end{equation}
where $\mathcal R$ contains gradient-penalty and weight-regularization terms. The detached AMP style reward enters PPO returns.

Detaching $\tilde x_t$ blocks PFM-loss gradients to the actor encoder; detaching $\tilde e_t$ blocks PPO gradients to PFM; and detached motion predictions block AMP-loss gradients to the motion head and HC backbone. Deployment retains the actor, including PFM, and removes the critic, simulator targets, and AMP discriminator.
\section{Experiments}

This section evaluates HC2Robot through the simulation protocol, deployable PFM prediction accuracy, comparative traversal and foothold-support results, and component ablations. We further assess zero-shot transfer on individual and mixed terrains in the real-world deployment.

\begin{figure}[!t]
\centering
\includegraphics[width=\columnwidth]{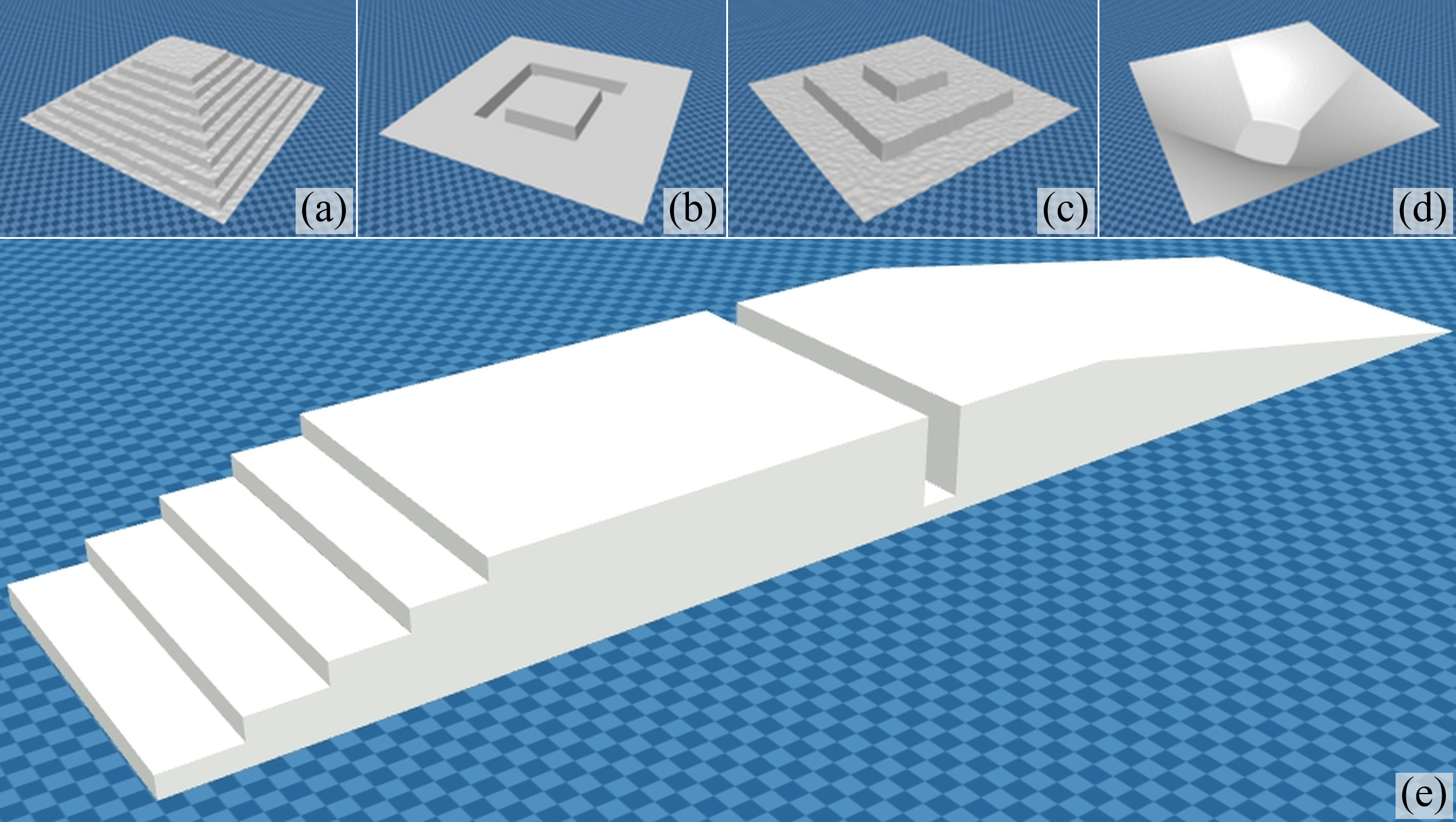}
\caption{Evaluation terrains: (a) stairs, (b) gaps, (c) high platforms, (d) slopes, and (e) a mixed course combining stairs, a gap, and a slope.}
\label{fig:terrain_overview}
\end{figure}

\begin{table}[!t]
\footnotesize
\centering
\caption{Training and mixed-course terrain parameters.}
\label{tab:terrain_settings}
\resizebox{\columnwidth}{!}{%
\begin{tabular}{lccc}
\toprule
\multicolumn{4}{c}{\textit{Training Terrains}} \\
\midrule
Terrain & \multicolumn{2}{c}{Attribute} & Range \\
\midrule
Stairs & \multicolumn{2}{c}{Step height (m)} & 0.05--0.23 \\
Gaps & \multicolumn{2}{c}{Gap width (m)} & 0.10--0.70 \\
High platforms & \multicolumn{2}{c}{Height (m)} & 0.05--0.45 \\
Slopes & \multicolumn{2}{c}{Slope (rise/run)} & 0.00--0.70 \\
\midrule
\multicolumn{4}{c}{\textit{Deployment Validation: Mix Terrain}} \\
\midrule
Difficulty & Stairs (m) & Gap (m) & Slope (${}^\circ$) \\
\midrule
Level 1 (Easy) & 0.090 & 0.175 & 10.0 \\
Level 2 (Moderate) & 0.160 & 0.325 & 14.7 \\
Level 3 (Higher) & 0.230 & 0.475 & 23.6 \\
\bottomrule
\end{tabular}
}
\end{table}

\subsection{Experimental Protocol}
All reported strategies were trained for 30,000 iterations with 4096 IsaacLab humanoid environments on an NVIDIA RTX 4090. Stairs, gaps, high platforms, and slopes randomize geometry and traversal direction at each iteration. All methods share the command distribution, terrain curriculum, and evaluation settings, including terrain configurations, termination criteria, and the SFR threshold; testing uses only the deployable actor and onboard observations. Simulation SR/SFR results summarize five seeds, each evaluated over 1000 trials per terrain class; means and standard deviations are computed across seeds. Fig.~\ref{fig:terrain_overview} shows the four training terrain classes and the mixed course, while Table~\ref{tab:terrain_settings} lists their training ranges and held-out mixed-terrain validation settings.

\begin{table}[!t]
\footnotesize
\centering
\caption{PFM prediction errors on held-out episodes. Mean $\pm$ standard deviation of per-seed mean errors.}
\label{tab:pfm_accuracy}
\setlength{\tabcolsep}{3pt}
\begin{tabular*}{\linewidth}{@{\extracolsep{\fill}}lccc@{}}
\toprule
Terrain & Time error & Position error & Support-ratio error \\
         & (steps)  & (m)         & (pp) \\
\midrule
Stairs         & \meanstd{0.73}{0.14} & \meanstd{0.018}{0.003} & \meanstd{2.76}{0.37} \\
Gaps           & \meanstd{0.91}{0.19} & \meanstd{0.021}{0.004} & \meanstd{1.19}{0.48} \\
High platforms & \meanstd{0.66}{0.11} & \meanstd{0.016}{0.002} & \meanstd{4.33}{0.24} \\
Slopes         & \meanstd{0.84}{0.17} & \meanstd{0.020}{0.003} & \meanstd{0.65}{0.41} \\
Overall        & \meanstd{0.79}{0.08} & \meanstd{0.019}{0.002} & \meanstd{2.23}{0.17} \\
\bottomrule
\end{tabular*}
\end{table}

\begin{table*}[!t]
\scriptsize
\centering
\caption{Simulation SR and SFR (\%). Values are mean $\pm$ standard deviation over five random seeds.}
\label{tab:sim_results}
\setlength{\tabcolsep}{1pt}
\resizebox{\textwidth}{!}{%
\begin{tabular}{lcccccccc}
\toprule
Method & \multicolumn{2}{c}{Stairs} & \multicolumn{2}{c}{Gaps} & \multicolumn{2}{c}{High Platforms} & \multicolumn{2}{c}{Slopes} \\
\cmidrule(lr){2-3}\cmidrule(lr){4-5}\cmidrule(lr){6-7}\cmidrule(lr){8-9}
 & SR & SFR & SR & SFR & SR & SFR & SR & SFR \\
\midrule
Hiking-in-the-Wild & \meanstd{68.10}{0.28\%} & \meanstd{84.86}{0.61\%} & \meanstd{58.96}{0.21\%} & \meanstd{87.04}{0.71\%} & \meanstd{76.14}{0.33\%} & \meanstd{85.66}{0.73\%} & \meanstd{70.54}{0.35\%} & \meanstd{81.92}{0.56\%} \\
ParkourFormer & \meanstd{94.98}{0.21\%} & \meanstd{96.26}{0.54\%} & \meanstd{95.20}{0.32\%} & \meanstd{96.33}{0.50\%} & \meanstd{91.18}{0.24\%} & \meanstd{96.63}{0.52\%} & \meanstd{95.32}{0.34\%} & \meanstd{89.77}{0.67\%} \\
HC2Robot (Ours) & \boldmeanstd{96.22}{0.40\%} & \boldmeanstd{98.88}{0.35\%} & \boldmeanstd{95.60}{0.27\%} & \boldmeanstd{99.13}{0.57\%} & \boldmeanstd{98.84}{0.44\%} & \boldmeanstd{98.68}{0.83\%} & \boldmeanstd{99.72}{0.21\%} & \boldmeanstd{95.08}{0.54\%} \\
\bottomrule
\end{tabular}
}
\end{table*}

\subsection{Metrics}
The primary task metric is success rate (SR), the fraction of episodes that complete the terrain segment without unsafe termination. Failures include falls, illegal non-foot contacts, leaving the valid terrain region, insufficient base height, and failure to make progress before the time limit.

Safe foothold rate (SFR) uses a geometric support proxy. Contact is detected when the maximum resultant foot force over the sensor history exceeds 1\,N. Each new contact is measured once after two consecutive contact steps. A yaw-aligned horizontal patch of $0.225\times0.10$\,m, centered 0.04\,m forward of the ankle, is sampled at 0.025\,m spacing, yielding 50 vertical terrain rays. For event $i$, let $h_{ij}$ denote sample $j$'s terrain height and $h_i^{\max}$ the highest finite hit. The evaluation support ratio and SFR are
\begin{equation}
\begin{aligned}
r_i^{\mathrm{eval}} &= \frac{1}{50}\sum_{j=1}^{50}\mathbf{1}[h_{ij}\text{ finite},\;h_i^{\max}-h_{ij}<0.025\,\mathrm{m}],\\
\mathrm{SFR} &= \frac{100\%}{N_c}\sum_{i=1}^{N_c}\mathbf{1}[r_i^{\mathrm{eval}}\geq0.75],
\end{aligned}
\label{eq:sfr_metric}
\end{equation}
where ray misses count as unsupported and an all-miss patch has zero support. $N_c$ pools measured events per seed and terrain class, including those recorded in failed trials up to failure detection; frozen environments contribute no further events. This ratio differs from the PFM supervision target in \eqref{eq:foothold_descriptor}. The horizontal patch ignores foot pitch and roll and can penalize continuous slopes; SFR therefore complements SR as a geometric proxy rather than a direct measure of physical contact or dynamic stability.

\subsection{PFM Prediction Accuracy}
To directly assess the foresight supplied to the actor, we evaluate PFM on held-out simulation episodes using only the deployable observation history. The point predictions in \eqref{eq:event_summary} are compared with simulator-derived targets in \eqref{eq:event_label} for offline scoring. Timing and position errors are $|\tilde\tau_f-\tau_f^*|$ and $\|\boldsymbol\mu_{p,f}-\boldsymbol p_f^*\|_2$, respectively; support-ratio error is $|\hat r_f-r_f^*|$, multiplied by 100 to express it in percentage points. For each seed and terrain class, errors are averaged over frame--foot samples with an observed future touchdown; right-censored samples are excluded. The Overall row averages the four terrain-class means within each seed. Table~\ref{tab:pfm_accuracy} reports overall mean errors of 0.79 control steps, 0.019\,m, and 2.23 percentage points for timing, position, and support ratio, respectively.

\subsection{Compared Methods}
We compare against \emph{Hiking-in-the-Wild} \cite{zhu2026hiking} and \emph{ParkourFormer} \cite{mai2026parkourformer}. Table~\ref{tab:model_size} reports model size and runtime.

\begin{figure}[!t]
\centering
\includegraphics[trim={5.01874pt 4.01499pt 40.14990pt 33.12367pt},clip,width=0.96338487\columnwidth]{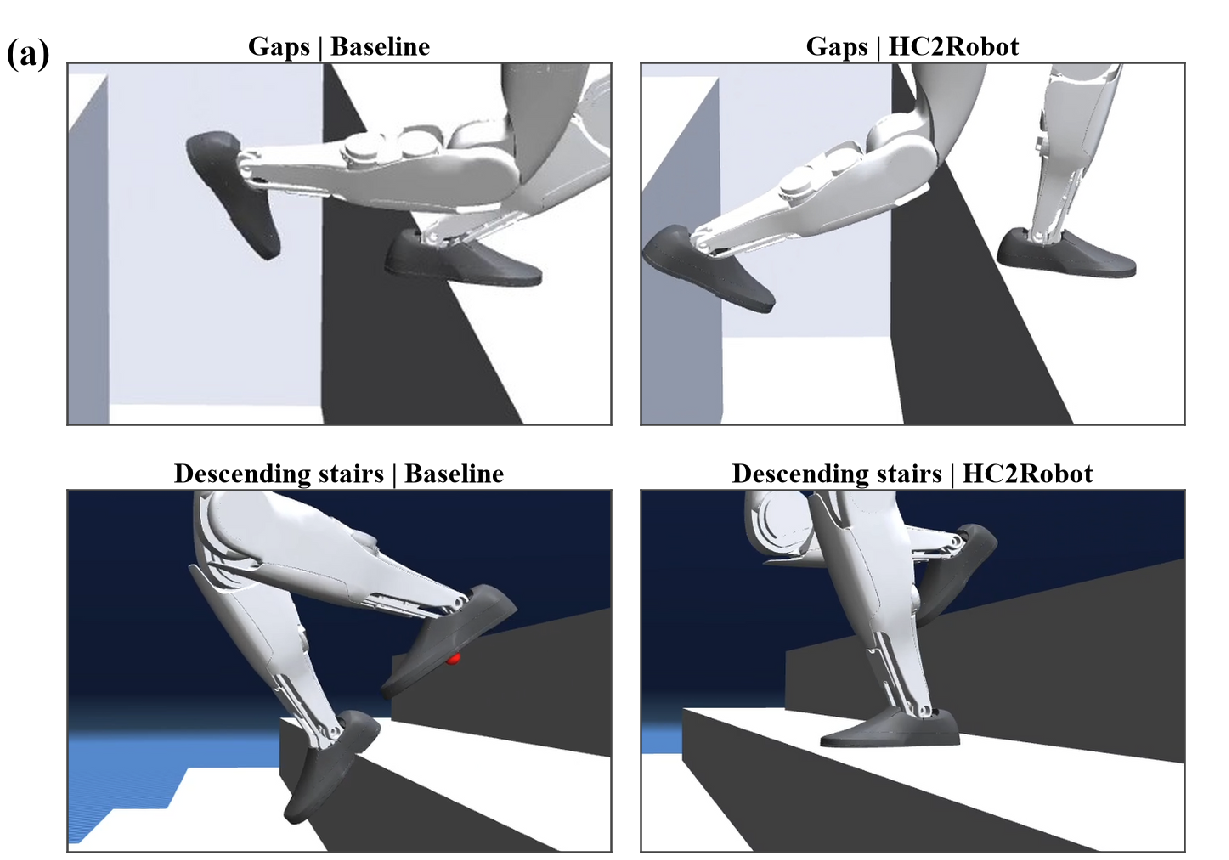}
\caption{Foot placement of the baseline and HC2Robot during gap crossing (top) and stair descent (bottom).}
\label{fig:foothold_comparison}
\end{figure}

\begin{figure}[!t]
\centering
\includegraphics[trim={17.06371pt 5.01874pt 12.04497pt 17.06371pt},clip,width=0.97573222\columnwidth]{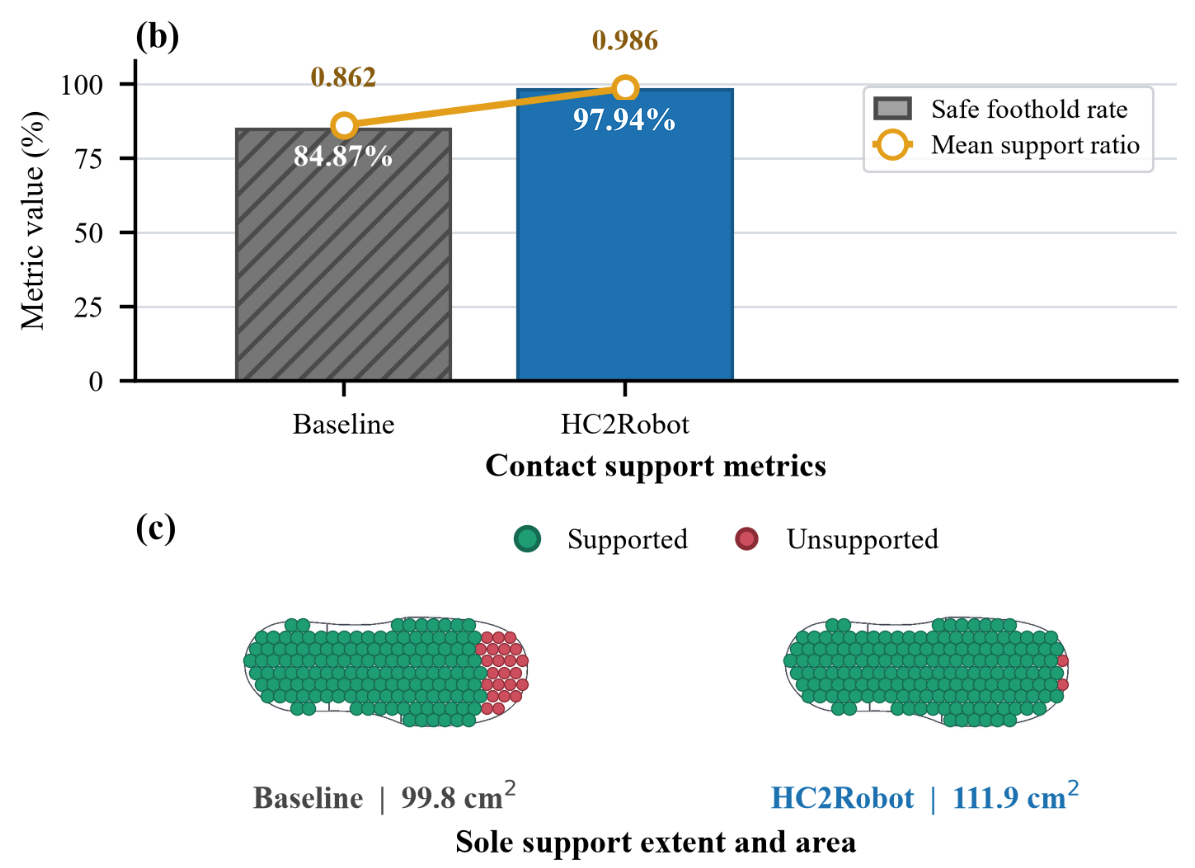}
\caption{Support vs. Hiking-in-the-Wild: (b) SFR (Table~\ref{tab:sim_results} mean) and mean support ratio; (c) schematic sole (green/red: supported/unsupported).}
\label{fig:foot_area}
\end{figure}

\subsection{Main Results}
Table~\ref{tab:sim_results} reports simulation performance across all terrain classes. HC2Robot attains the highest SR and SFR on every terrain. Its unweighted mean SR and SFR across the four terrain classes are 97.60\% and 97.94\%, respectively, compared with 94.17\% and 94.75\% for ParkourFormer. The largest terrain-specific gains over ParkourFormer are 7.66 SR points on high platforms and 5.31 SFR points on slopes; SFR on gaps also improves by 2.80 points. Figs.~\ref{fig:foothold_comparison} and \ref{fig:foot_area} illustrate foot placement and geometric support, complementing the aggregate SR/SFR results.

\begin{table}[!t]
\footnotesize
\centering
\caption{Model complexity, training duration, and inference latency.}
\label{tab:model_size}
\setlength{\tabcolsep}{4pt}
\resizebox{\columnwidth}{!}{%
\begin{tabular}{lcccc}
\toprule
Model & Params. (M) & FLOPs (M) & Train (h) & Infer. (ms) \\
\midrule
\textbf{Hiking-in-the-Wild} & 3.58 & 3.89 & 50 & 0.28 \\
\textbf{ParkourFormer} & 3.74 & 7.94 & 89 & 0.53 \\
\textbf{HC2Robot (Ours)} & 3.67 & 9.77 & 77 & 0.40 \\
\bottomrule
\end{tabular}
}
\end{table}

\subsection{Auxiliary Diagnostics and Ablations}

We evaluate the contributions of the main components through ablations. \emph{w/o Hyper-Connection} replaces HC with a standard residual Transformer; \emph{w/o PFM} removes the estimator and its adapter; \emph{Auxiliary-only} disables touchdown-event supervision and removes the corresponding event features from policy fusion while retaining the one-step base-velocity and center-of-mass predictions; and \emph{w/o AMP} removes the adversarial motion prior. All variants use the same base deployable observations, critic-side privileged inputs, training budget, and evaluation protocol.

The \emph{Fixed-horizon} control predicts the same six per-foot position and support attributes at $t+h_0$, with $h_0=60$ control periods, instead of at the next touchdown, and omits the event-time loss and event-time input. The offset $h_0$ is fixed across terrains and random seeds. The control retains HC, the one-step auxiliary heads, adapter, late fusion, training budget, and evaluation settings. This is a composite fixed-offset comparison that changes both target alignment and event-time information; it does not isolate either factor separately.

\begin{table}[H]
\scriptsize
\centering
\caption{Component ablations for SR and SFR (\%). Plat.\ denotes high platforms; Mean averages the four terrains.}
\label{tab:ablation}
\setlength{\tabcolsep}{1.4pt}
\renewcommand{\arraystretch}{0.92}
\resizebox{\columnwidth}{!}{%
\begin{tabular}{lccccc}
\toprule
Method & Stairs & Gaps & Plat. & Slopes & Mean \\
\midrule
\multicolumn{6}{l}{\emph{SR (\%)} $\uparrow$} \\
w/o PFM & \meanstd{88.12}{0.32\%} & \meanstd{90.34}{0.42\%} & \meanstd{97.88}{0.40\%} & \meanstd{93.66}{0.49\%} & \meanstd{92.50}{0.21\%} \\
Auxiliary-only & \meanstd{90.20}{0.30\%} & \meanstd{92.76}{0.42\%} & \meanstd{91.20}{0.33\%} & \meanstd{91.84}{0.29\%} & \meanstd{91.50}{0.10\%} \\
w/o Hyper-Connection & \meanstd{93.86}{0.25\%} & \meanstd{95.26}{0.37\%} & \meanstd{98.12}{0.42\%} & \meanstd{75.16}{0.38\%} & \meanstd{90.60}{0.08\%} \\
w/o AMP & \meanstd{92.24}{0.33\%} & \meanstd{88.54}{0.28\%} & \meanstd{88.12}{0.33\%} & \meanstd{86.34}{0.45\%} & \meanstd{88.81}{0.17\%} \\
Fixed-horizon & \meanstd{91.04}{0.21\%} & \meanstd{92.86}{0.13\%} & \meanstd{89.76}{0.36\%} & \meanstd{93.96}{0.30\%} & \meanstd{91.91}{0.10\%} \\
Ours & \boldmeanstd{96.22}{0.40\%} & \boldmeanstd{95.60}{0.27\%} & \boldmeanstd{98.84}{0.44\%} & \boldmeanstd{99.72}{0.21\%} & \boldmeanstd{97.60}{0.19\%} \\
\midrule
\multicolumn{6}{l}{\emph{SFR (\%)} $\uparrow$} \\
w/o PFM & \meanstd{93.53}{0.45\%} & \meanstd{94.08}{0.73\%} & \meanstd{95.37}{0.61\%} & \meanstd{90.85}{0.77\%} & \meanstd{93.46}{0.24\%} \\
Auxiliary-only & \meanstd{92.56}{0.50\%} & \meanstd{90.18}{0.71\%} & \meanstd{92.72}{0.66\%} & \meanstd{84.62}{0.45\%} & \meanstd{90.02}{0.22\%} \\
w/o Hyper-Connection & \meanstd{94.47}{0.40\%} & \meanstd{93.55}{0.84\%} & \meanstd{95.25}{0.68\%} & \meanstd{76.72}{0.78\%} & \meanstd{90.00}{0.22\%} \\
w/o AMP & \meanstd{91.17}{0.37\%} & \meanstd{94.98}{0.48\%} & \meanstd{92.20}{0.61\%} & \meanstd{69.81}{0.88\%} & \meanstd{87.04}{0.24\%} \\
Fixed-horizon & \meanstd{96.25}{0.53\%} & \meanstd{93.06}{0.61\%} & \meanstd{95.52}{0.73\%} & \meanstd{91.08}{0.68\%} & \meanstd{93.98}{0.15\%} \\
Ours & \boldmeanstd{98.88}{0.35\%} & \boldmeanstd{99.13}{0.57\%} & \boldmeanstd{98.68}{0.83\%} & \boldmeanstd{95.08}{0.54\%} & \boldmeanstd{97.94}{0.31\%} \\
\bottomrule
\end{tabular}
}
\end{table}

\paragraph{Component Ablation Results.}

Table~\ref{tab:ablation} reports SR and SFR for the completed variants. Removing PFM reduces mean SR/SFR from 97.60\%/97.94\% to 92.50\%/93.46\%, supporting the benefit of the complete PFM pathway for traversal and geometric foothold support. The Auxiliary-only variant obtains 91.50\%/90.02\%. Its lower performance than the full model indicates that the supervised touchdown-event representation contributes beyond the shared auxiliary predictions; this comparison does not attribute the difference to a single event target. The w/o Hyper-Connection variant is especially weak on slopes (75.16\% SR, 76.72\% SFR). Removing AMP yields the lowest aggregate means (88.81\% SR, 87.04\% SFR) and the lowest slope SFR (69.81\%).

HC2Robot also exceeds Fixed-horizon in SR and SFR on every terrain, with the largest gains of 9.08 SR points on high platforms and 6.07 SFR points on gaps. This result shows that the complete event-based design outperforms the evaluated fixed-offset, no-event-time alternative.

\begin{figure}[H]
\centering
\includegraphics[trim={9.15418pt 6.50428pt 2.64989pt 6.74518pt},clip,width=0.95333333\columnwidth]{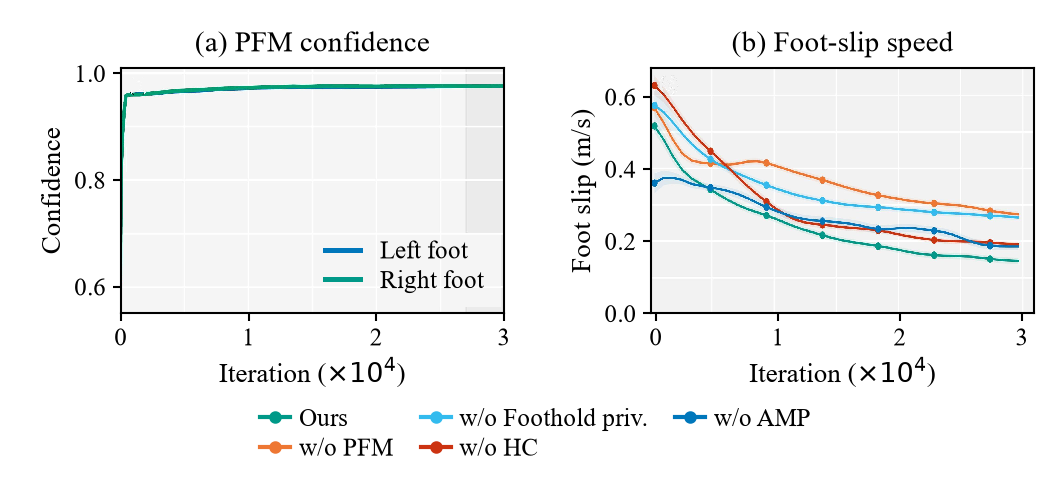}
\caption{Training traces: (a) PFM confidence; (b) foot-slip speed. Binned, smoothed single-run curves; shading: within-bin temporal SD.}
\label{fig:training_diagnostics}
\end{figure}

\paragraph{Training Dynamics.}
Fig.~\ref{fig:training_diagnostics} shows similar left/right confidence and the lowest terminal foot-slip speed for the full model among the displayed single-run traces. The w/o Foothold priv. trace is an earlier supervision-removal diagnostic, distinct from the Auxiliary-only control in Table~\ref{tab:ablation}.

\subsection{Real-World Deployment}
\label{sec:realworld}

We deploy HC2Robot zero-shot on the Unitree G1, with onboard inference at 50\,Hz using only proprioception and an egocentric depth camera.

\begin{table}[H]
\scriptsize
\centering
\caption{Real-world success rates. Each stairs trial includes one ascent and descent.}
\label{tab:real_results}
\setlength{\tabcolsep}{2pt}
\renewcommand{\arraystretch}{0.95}
\begin{tabular*}{\columnwidth}{@{\extracolsep{\fill}}llc@{}}
\toprule
\textbf{Terrain} & \textbf{Settings} & \textbf{Success Rate} \\
\midrule
\multicolumn{3}{c}{\textit{Mixed-Terrain Settings}} \\
\midrule
Mix-Level1 & Stairs / Gaps / Slopes &  100\%(20/20) \\
Mix-Level2 & Stairs / Gaps / Slopes &  95\%(19/20) \\
Mix-Level3 & Stairs / Gaps / Slopes &  85\%(17/20) \\
\midrule
\multicolumn{3}{c}{\textit{Training-Terrain Settings}} \\
\midrule
Stairs & 0.16\,m &  100\%(20/20) \\
Gaps & 0.40\,m &  100\%(20/20) \\
Platforms & 0.40\,m &  95\%(19/20) \\
Slopes & 0.50 rise/run &  100\%(20/20) \\
\bottomrule
\end{tabular*}
\end{table}

\begin{figure}[H]
\centering
\includegraphics[width=1\columnwidth]{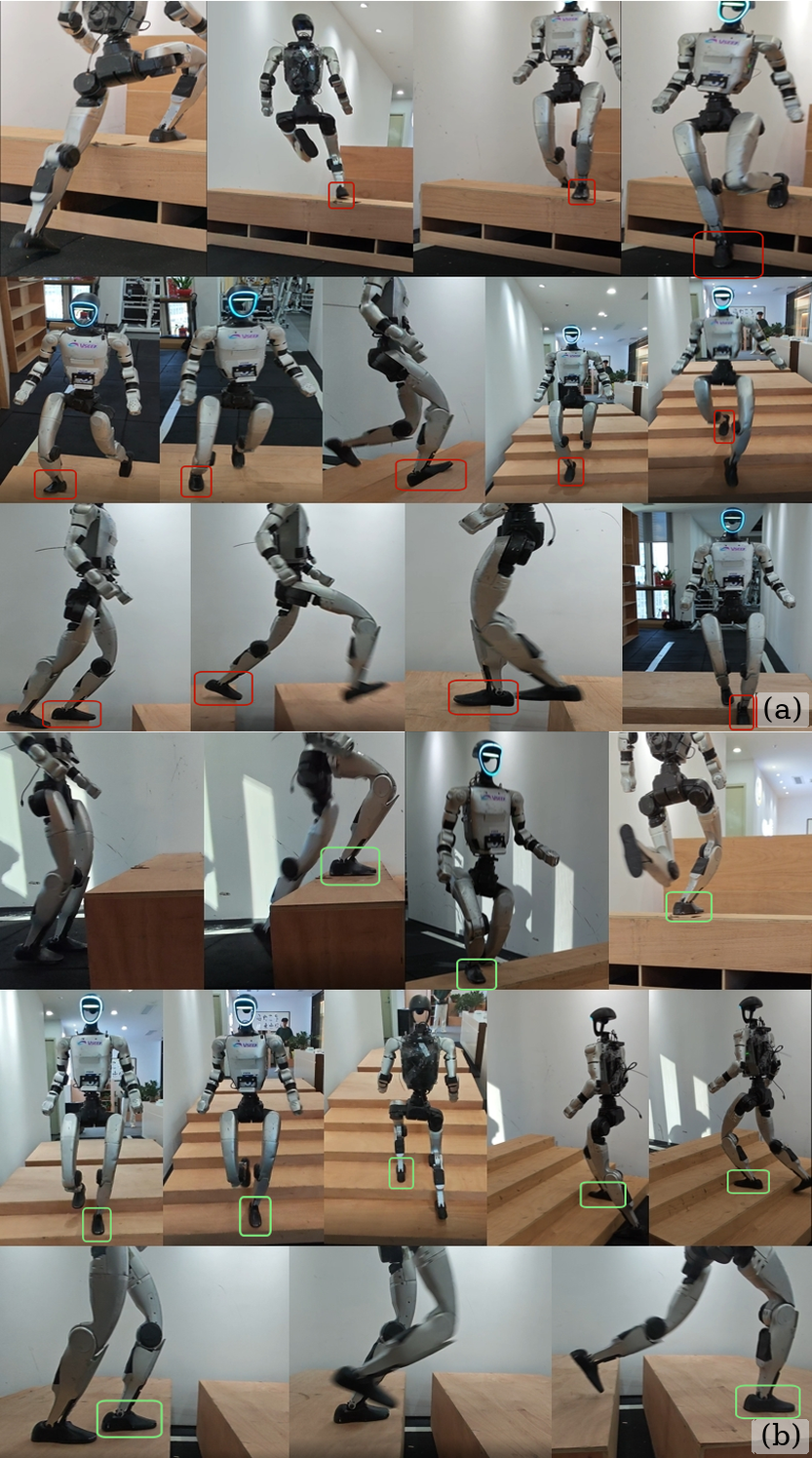}
\caption{Unitree G1 deployment snapshots: (a) baseline and (b) HC2Robot.}
\label{fig:realworld}
\end{figure}

We evaluate physical realizations of the training terrain classes and mixed courses combining stairs, gaps, and slopes at three difficulty levels. Single-terrain trials test sim-to-real transfer; mixed courses test compositional generalization to terrain arrangements and transitions not encountered during training, rather than new terrain primitives. Each setting uses 20 trials under the simulation failure criteria; Table~\ref{tab:real_results} reports complete-trial success rates, and Fig.~\ref{fig:realworld} shows traversal examples.

\section{Conclusion}

We presented HC2Robot, combining HC multi-stream routing and PFM touchdown-event foresight under asymmetric actor--critic training. PFM learns from simulator supervision to predict each foot's next touchdown timing, location, and support quality, while deployment uses only proprioception and depth. Held-out evaluation quantifies touchdown timing, position, and support-ratio errors. Across four simulated terrain classes, HC2Robot achieves mean SR/SFR of 97.60\%/97.94\%, outperforming the evaluated baselines. Ablations assess PFM, HC, and the motion prior; Auxiliary-only supports the benefit of supervised touchdown-event features beyond shared one-step kinematic predictions. The fixed-horizon comparison evaluates the complete event-based design against a fixed-offset, no-event-time alternative. Zero-shot Unitree G1 trials demonstrate transfer to individual obstacles and mixed courses.

\bibliographystyle{IEEEtranBST/IEEEtran}
\bibliography{references}
\end{document}